# ORICF - Open Robotics Inference and Control Framework⋆

Andrés Meseguer Valenzuela[1,*,†], Luís Miguel Bartolín Arnau[1,†]

[1] *Instituto Tecnológico de Informática (ITI), Paterna, Spain*

**Abstract**

Recent advances in artificial intelligence (AI) have enabled effective perception and language models for robots, but their deployment remains computationally expensive, increasing latency and energy use. This work presents the Open Robotics Inference and Control Framework (ORICF), a modular, declarative, and model-agnostic platform for composing multimodal robotic inference pipelines. ORICF integrates input/output (I/O) adapters, pluggable inference back ends, and post-processing logic, while lightweight YAML specifications allow models, hardware targets, and data channels to be changed without code modification. The framework also supports edge offloading, i.e., executing inference on nearby external computers instead of onboard the robot. ORICF is evaluated on a mobile robot that answers spoken queries about people detected in its camera stream by combining automatic speech recognition (ASR), a large language model (LLM), and a convolutional neural network (CNN) detector through Robot Operating System 2 (ROS2). Compared with onboard execution, ORICF-based edge deployment reduces robot-side compute utilization by up to 83.16% and estimated energy consumption by 65.8%, while preserving modularity and reproducibility.

**Keywords**

AI for robotic systems, Machine Learning, Physical AI [1]

## 1. Introduction

In recent years, machine learning models for robotics have advanced rapidly, particularly in perception, language understanding, and control. Although many are compatible with the Robot Operating System (ROS), their practical adoption still demands substantial integration effort, limiting both research use and deployable applications. In addition, running advanced models directly on embedded robotic platforms increases latency, energy consumption, and computational load, reducing autonomy. This motivates not only offloading inference to nearby edge resources, but also software frameworks that make such distributed deployments easier to configure, reproduce, and adapt. At the same time, Artificial Intelligence (AI) resources for robotics have expanded, including optimized runtimes such as Open Neural Network Exchange (ONNX) Runtime and repositories such as Hugging Face and OpenMMLab. However, these tools are typically provided as isolated components, requiring users to adapt models, pre-processing, communication interfaces, and inference pipelines for robotic use. They also offer limited support for real-world streams such as ROS2 topics or Real-Time Streaming Protocol (RTSP) feeds. Consequently, despite the availability of multimodal and Large Language Model (LLM)-based tools, the lack of unified protocol-aware frameworks continues to hinder practical deployment.

To address this gap, this work introduces the Open Robotics Inference and Control Framework (ORICF), a modular, declarative, and model-agnostic platform for AI-driven robotics. The

[1*] Corresponding author.
[†] These authors contributed equally.
ameseguer@iti.es (A. Meseguer Valenzuela); lbartolin@iti.es (L. M. Bartolín Arnau)
0000-0002-2328-6422 (A. Meseguer Valenzuela); 0009-0007-0904-6653 (L. M. Bartolín Arnau)

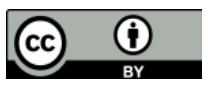

contribution of this paper is threefold. First, ORICF provides a unified architecture that decouples input/output channels, inference back ends, and post-processing logic, reducing the effort required to integrate AI models into robotic systems. Second, it enables declarative multimodal pipelines in which heterogeneous models, such as vision, speech, and language models, can be chained and reconfigured through lightweight YAML files. Third, it supports flexible deployment across onboard and edge resources, allowing computationally demanding inference tasks to be offloaded without rewriting the robotic application.

ORICF comprises three modules: input/output, inference, and post-processing. Applications are deployed declaratively through Yet Another Markup Language (YAML) configuration files, avoiding complex integration and improving reproducibility across platforms. The framework also supports multiple models within one pipeline, enabling multimodal chaining between perception and language components. In this work, a demonstration combines an LLM with a convolutional neural network (CNN): the former processes voice commands, while the latter detects people in a camera stream. This pipeline enables a robot running ROS2 to answer natural-language questions about visible people, showing how ORICF unifies perception, reasoning, and robotic communication in a reproducible deployment.

The experiments demonstrate that an ORICF-based edge deployment reduces the robot processing load by 83.16% and estimated energy consumption by 65.8% compared with onboard execution. These results show that ORICF can orchestrate multimodal robotic inference while enabling practical offloading without sacrificing modularity or reproducibility. The remainder of this article is organized as follows: Section II reviews the state of the art and compares ORICF with existing solutions; Section III presents the proposed architecture; Section IV details the experimental evaluation and results; and Section V concludes the paper, outlining future directions for improvement and extension.

## 2. State of the Art

Recent advancements in robotics have increasingly relied on cloud and edge computing for perception and natural language tasks [5], [6]. Frameworks like FogROS2 facilitate the offloading of computationally heavy tasks to external resources to address latency, energy, and cost trade-offs [1]–[4], [15], but they generally assume the application is already implemented as ROS2 nodes [3], [4]. Concurrently, Large Language Models (LLMs) are being integrated into robotics for high-level reasoning and policy generation [7], [10]–[12] through tools like PaLM-E [9], Code-as-Policies [8], ROS-LLM [13], and MCP servers [14]. Despite these developments, existing tools lack unified, declarative orchestration layers that can chain heterogeneous models (vision, ASR, LLM) and standardized I/O across different runtimes without requiring pipeline recoding [12]. As detailed in Table I, ORICF addresses this gap by providing a model-agnostic orchestration layer that supports multimodal chaining across heterogeneous models and I/O sources.

**Table 1**
Comparison of Robotic AI Integration Frameworks

| Solution | Applicable model types | Data I/O | Model-to-model feedback | Implementation | Edge / Cloud support |
|---|---|---|---|---|---|
| ORICF (presented) | Model agnostic | Multimodal | Yes | Python and C++ | Yes |
| ROS-LLM | LLM + VLM | Text and video | Limited | ROS-native | Limited |
| ROS-MCP server | LLM | Text | Limited | Python | Limited |

## 3. Architecture

ORICF is built as an orchestration layer between robotic data sources, inference back ends, and application-level outputs, providing a flexible alternative to monolithic robotics software stacks. Rather than relying on rigid pipelines, it follows a plug-and-play approach in which models and data channels are combined declaratively, reducing development time and deployment complexity. The framework is implemented mainly in Python for compatibility with machine learning libraries, while ROS2 communication is handled through rclpy. It also supports optimized runtimes such as OpenVINO and ONNX Runtime, with optional C++ bindings for real-time or embedded optimization.

The framework is organized into three modules. The input–output module abstracts data acquisition and publication across heterogeneous sources, including ROS2 topics, RTSP streams, audio devices, and text channels. The inference module provides a uniform interface to machine learning back ends, allowing perception, speech, and language models to be exchanged or combined without modifying the pipeline. The post-processing module adds application logic, such as label mapping, object counting, image annotation, or symbolic outputs for downstream models. Deployment relies on lightweight YAML files specifying the model, I/O topics, inference device, runtime, and semantic labels. Thus, users can switch data sources, redirect outputs, or move inference to an edge server through configuration alone.

ORICF interconnects multiple models by exposing outputs as configurable data products that can be consumed by downstream modules or published through ROS2 topics. This enables multimodal chaining between perception, speech, and language components. In the demonstration, a CNN detects people in a camera stream, post-processing converts detections into a person count, and an LLM combines this information with spoken commands to generate a natural-language answer. This demonstration shows that ORICF supports flexible, reproducible, and practical deployment of multimodal robotic pipelines.

## 4. Evaluation

The objective of the evaluation is to validate ORICF as an integration and deployment layer for multimodal robotic inference, and to quantify the effect of onboard and edge execution on computational load and estimated energy consumption. The selected use case focuses on human–robot interaction, in which a mobile robot answers a spoken query in natural language by reporting the number of people detected in its field of view. This scenario combines perception, language understanding, and real-time response, thereby exercising both the orchestration capabilities of the framework and the resource impact of distributed execution.

In the experiment, the interaction begins with a spoken command. The audio input is transcribed to text through an automatic speech recognition pipeline, after which the resulting text is processed by a large language model. In parallel, a convolutional neural network, namely Person-detection-0200, detects humans in red-green-blue frames acquired from the robot camera and streamed through the ROS2 topic camera/image_raw. A custom post-processing module counts the detected persons and publishes the result on the topic /human_counter. This information is then incorporated into the language model response, enabling the robot to answer queries such as “How many people do you see?”. The experiment highlights ORICF’s modularity, since heterogeneous models are orchestrated within the same pipeline without pipeline recoding.

The experimental infrastructure comprised both onboard and edge resources. The robotic platform was an X100 mobile robot manufactured by XMachines, equipped with an NVIDIA Jetson Orin NX embedded computer with 8 GB of RAM and an Intel RealSense D455 RGB camera. The robot software stack ran Ubuntu 22.04 and ROS2 Humble. External inference tasks were executed on a Proxmox-hosted edge server with a 12-core processor at 2.1 GHz and 16 GB of RAM, also running Ubuntu 22.04 and ROS2 Humble. Within this infrastructure, Person-detection-0200 was used for visual perception, while TinyLlama-1.1B-Chat-v1.0 was employed as the language model. ORICF coordinated execution across both environments, enabling comparison between onboard processing and ORICF-based edge offloading. Computational load was measured across three operating phases: (i) baseline system activity without inference, (ii) activation of the person detection CNN only, and (iii) full pipeline execution with both CNN and LLM. Measurements were performed using vmstat on both devices. The results show that onboard execution approaches saturation when the LLM is included, whereas edge offloading substantially reduces the burden on the embedded system.

Energy consumption was estimated using a theoretical model that relates processor utilization to instantaneous power draw. Specifically, the power consumed by the robot was modeled in (1), where u denotes the average processing utilization, $P_{idle}$ is the power consumption of the robot when the processor is idle, and $P_{full}$ represents power consumption under full load. To quantify the impact of offloading inference tasks to the edge, the relative reduction in (2) was used, where $P_{CNN+LLM}$ corresponds to the estimated power when the robot executes both the CNN and LLM locally, and $P_{BASE}$ denotes the power when these tasks are offloaded and the robot only runs its base functionalities. Applying this model to the experimental data indicates a clear reduction in estimated robot-side energy consumption when computation is shifted to the edge.

$$P_u = P_{idle} + u \cdot (P_{full} - P_{idle}), \tag{1}$$

$$R = 1 - \frac{P_{BASE}}{P_{(CNN+LLM)}} \tag{2}$$

For the Jetson Orin NX mounted on the robot, a baseline idle power of 5 W ($P_{idle}$) and a maximum load power of 25 W ($P_{full}$) were considered, consistent with vendor specifications. Using experimental processing utilization data, the median utilization while running both CNN and LLM locally was 95%, which yields an estimated consumption of 24 W ($P_{CNN+LLM}$). In contrast, when inference is offloaded to the edge, the robot only executes its basic functions, with a median utilization of 16%, resulting in 8.2 W base consumption ($P_{BASE}$). Substituting these values in (2), a relative reduction of 0.658 is obtained, corresponding to a 65.8% decrease in estimated energy consumption. This theoretical estimation indicates that offloading inference alleviates computational load and may contribute to energy savings, which are critical for extending the operational autonomy of mobile robots.

**Table 2**
Robot and edge processing utilization across execution modes

| Value | Robot base | Robot CNN | Robot CNN + LLM | Edge base | Edge CNN | Edge CNN + LLM |
|---|---|---|---|---|---|---|
| Average (%) | 18.65 | 74.95 | 88.45 | 7.80 | 29.40 | 42.05 |
| Median (%) | 16.00 | 76.00 | 95.00 | 7.00 | 30.00 | 43.00 |
| Variance | 27.53 | 42.25 | 400.15 | 17.66 | 95.24 | 495.05 |
| Standard Deviation (%) | 5.25 | 6.50 | 20.00 | 4.20 | 9.76 | 22.25 |

Table 2 summarizes the processing utilization statistics obtained for both the robot and the edge server, considering three operational modes: baseline, CNN inference, and CNN+LLM inference. Results show that the robot reaches a median processing utilization of 95% when running CNN+LLM locally, while this drops to 16% when computation is delegated to the edge, corresponding to an 83.16% reduction in robot-side computing load.

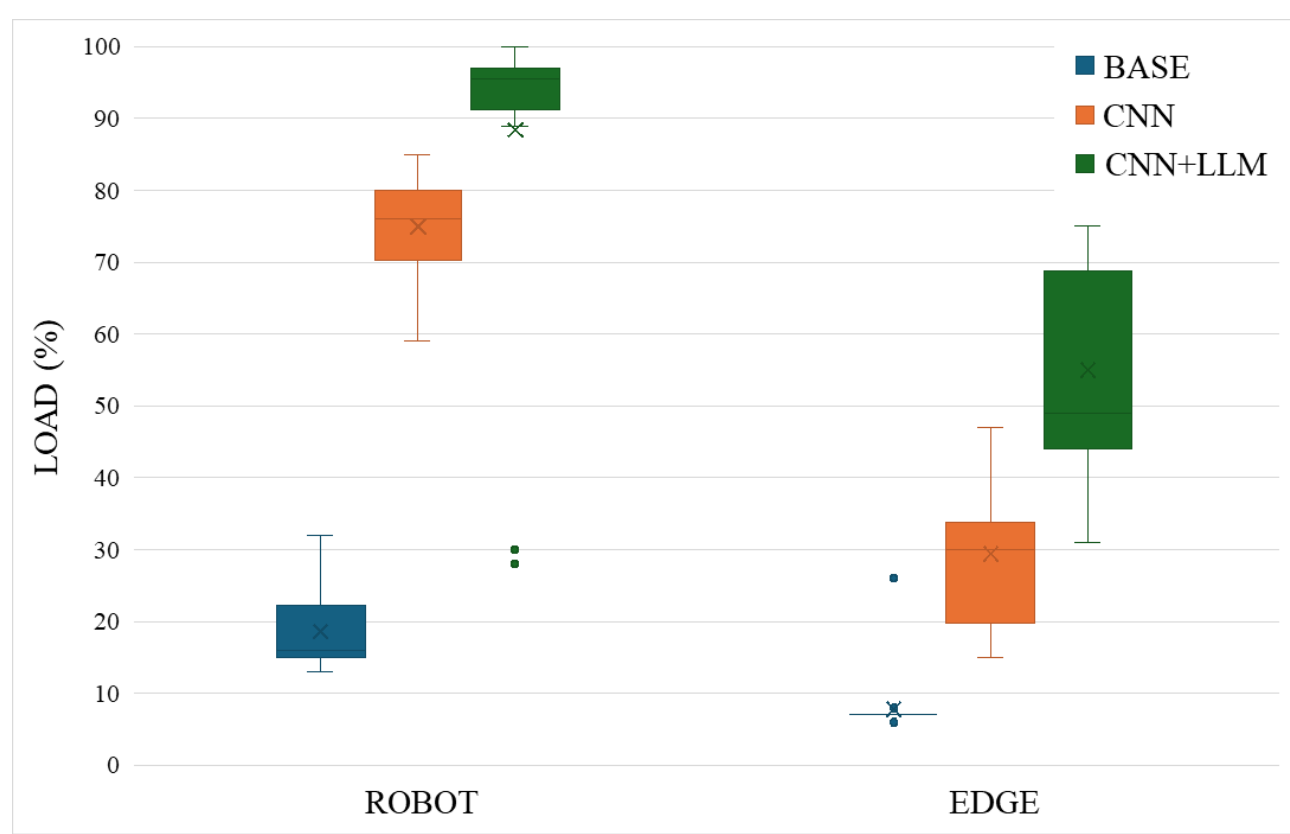

**Figure 1:** Processing utilization comparison under onboard and edge execution

Furthermore, workload variability shows that local execution of the CNN and LLM introduces high dispersion (σ=20), which may affect real-time responsiveness. Conversely, edge offloading keeps the embedded system well below saturation, reducing both computational demand and resource-use variability. These results are shown in Fig. 1. Overall, the experiments indicate that ORICF can orchestrate heterogeneous models in a distributed robotic pipeline while enabling the benefits of edge execution. However, the evaluation does not isolate ORICF’s overhead against a manually integrated ROS2 pipeline, nor does it assess network degradation, security mechanisms, or fault tolerance, which remain relevant aspects for future deployments.

## 5. Conclusions

ORICF provides a modular, declarative, and model-agnostic platform for deploying multimodal AI pipelines in robotics through lightweight YAML configurations. By separating input/output handling, inference back ends, and post-processing logic, the framework facilitates the composition of heterogeneous models across onboard and edge resources without rewriting the robotic application. The experimental evaluation shows that ORICF-based edge deployment reduces robot-side processing load by 83.16% and estimated energy consumption by 65.8%, while lowering computational variability. These results support ORICF as an integration and deployment layer for resource-constrained robotic systems, with the efficiency gains mainly associated with edge execution. Future work will extend the evaluation to additional platforms, tasks, and network conditions, and will address framework overhead, security, and fault tolerance in real-world deployments.

## Acknowledgements

Research activities supported by Red de Excelencia Cervera para la Investigación en Tecnologías Avanzadas para la Cognición, simulación Táctica y entrenamiento adaptativo en escenarios críticos a través de Interfaces Extendidas (CER-20251027) funded by Ministerio de Ciencia, Innovación y Universidades a través del Centro para el Desarrollo Tecnológico y la Innovación (CDTI).

## Declaration on Generative AI

During the preparation of this work, the authors used ChatGPT5.2 for Grammar and spelling check. After using this tool, the authors reviewed and edited the content as needed and take full responsibility for the publication's content.